\documentclass[journal]{IEEEtran}
\usepackage{amsmath,amsfonts}
\usepackage{algorithmic}
\usepackage{algorithm}
\usepackage{array}
\usepackage[caption=false,font=normalsize,labelfont=sf,textfont=sf]{subfig}
\usepackage{textcomp}
\usepackage{stfloats}
\usepackage{multirow}
\usepackage{url}
\usepackage{verbatim}
\usepackage{graphicx}
\usepackage{cite}
\usepackage{wasysym}
\begin{document}

\title{The Complexity of Extreme Climate Events on \\the New Zealand's Kiwifruit Industry}

\author{
FA128 Project Team (Pre-print v0.8 2025-08-04)\\ 
\IEEEauthorblockN{Boyuan Zheng\IEEEauthorrefmark{1}, Victor W. Chu\IEEEauthorrefmark{1}, Zhidong Li\IEEEauthorrefmark{1}\\
Evan Webster\IEEEauthorrefmark{2}, Ashley Rootsey\IEEEauthorrefmark{2}\IEEEauthorrefmark{3}} \\   
\IEEEauthorrefmark{1}Data Science Institute, University of Technology Sydney \\
\IEEEauthorrefmark{2}Yamaha Agriculture \& \IEEEauthorrefmark{3}Food Agility CRC
}

%\thanks{This paper was produced by the IEEE Publication Technology Group. They are in Piscataway, NJ.}% <-this % stops a space
%\thanks{Manuscript received April 19, 2021; revised August 16, 2021.}}



\maketitle

\begin{abstract}
% outlier results, extreme events, agriculture yield
Climate change has intensified the frequency and severity of extreme weather events, presenting unprecedented challenges to the agricultural industry worldwide. In this investigation, we focus on kiwifruit farming in New Zealand. We propose to examine the impacts of climate-induced extreme events, specifically frost, drought, extreme rainfall, and heatwave, on kiwifruit harvest yields. These four events were selected due to their significant impacts on crop productivity and their prevalence as recorded by climate monitoring institutions in the country. We employed Isolation Forest, an unsupervised anomaly detection method, to analyse climate history and recorded extreme events, alongside with kiwifruit yields. Our analysis reveals considerable variability in how different types of extreme event affect kiwifruit yields underscoring notable discrepancies between climatic extremes and individual farm's yield outcomes. Additionally, our study highlights critical limitations of current anomaly detection approaches, particularly in accurately identifying events such as frost. These findings emphasise the need for integrating supplementary features like farm management strategies with climate adaptation practices. Our further investigation will employ ensemble methods that consolidate nearby farms' yield data and regional climate station features to reduce variance, thereby enhancing the accuracy and reliability of extreme event detection and the formulation of response strategies.
\end{abstract}


\section{Introduction}
% climate change and its impact on extreme events (EEs): freq and intense
Climate change is triggering notable shifts in global weather patterns resulting in an increased frequency and intensity of adverse climate events. Rising global temperatures have altered atmospheric conditions intensifying the occurrences of heatwave, drought, and heavy precipitation events. Recent studies, such as \cite{chen2018projected} and \cite{gaitan2020machine}, highlighted that even modest global temperature rises can substantially amplify these extreme events. Specifically, differences between warming scenarios of 1.5°C and 2.0°C above pre-industrial levels, as projected by the IPCC \cite{IPCC2018}, have shown significant variability in projected consequences. Historical climate records provide further evidence to the occurrence of these relations documenting a clear increase in temperature extremes and heavy rainfall episodes \cite{powell2016measuring}. As such climatic shifts intensify, having a comprehensive understanding of these extreme events is imperative for the development of effective mitigation and adaptation strategies.

% climate Change and extreme events impact on Kiwi industry
The kiwifruit industry has already been experiencing significant disruptions due to these extreme events. Rising temperatures, altered rainfall patterns, and increased frequency of extreme weather events have posed substantial challenges to the farmers. Warmer autumn and winter conditions are reducing winter chilling hours --- something crucial for kiwifruit budbreak and flower production which directly impacts both quantity and quality of yield --- has prompted the migration of cultivation practices toward higher latitudes. Furthermore, events such as hailstorms, severe winds, and intense rainfall are becoming increasingly common resulting in notable financial losses as documented by \cite{zespriannual}. Prolonged drought conditions and elevated evapotranspiration further heighten concerns around water availability prompting the need for improved irrigation practices and climate-resilient orchard management. As climate extremes grow more severe, proactive adaptation strategies will be essential for maintaining long-term viability and resilience within kiwifruit production.

% outlier detection for extreme events, limitation
A critical step towards assessing the impact of extreme events on kiwifruit yields involves accurately identifying and quantifying their occurrences, where national weather services leverage standardised indices such as SPI or percentage thresholds to detect such events. While such detection methods may not directly affect farmers' short-term day-to-day decisions, they are critical for analysis on modelling long-term yield risks and developing adaptation strategies. In this context, anomaly detection techniques play a key role though each of them offers distinct benefits but also limitations. Functional outlier detection methods, discussed in \cite{hael2020identifying}, utilise graphical tools such as rainbow plots to visualise anomalies yet often falter in high-dimensional scenarios and rely heavily on expert interpretation. Alternatively, multivariate anomaly detection approaches, which consider interactions among environmental variables, typically enhance accuracy though necessitate extensive pre-processing to handle seasonal and correlated effects \cite{flach2017multivariate}. Density-based clustering techniques like DBSCAN can effectively detect grouped anomalies but exhibit high sensitivity to parameter selection and may under-perform with irregular temporal patterns \cite{aslan2022detection}. Compared to these methods, Isolation Forest (IF) provides a computationally efficient and scalable alternative, adept at handling large and high-dimensional datasets due to its recursive partitioning approach \cite{bara2024anomaly}.

% our paper's contribution: climate limited impact on yield (management features); variance in influence of EEs; discrepancy and alignment; 
In this study, we examine the complex relationship between extreme climate events and kiwifruit yield variability in New Zealand, recognising the limitations of climate data alone in explaining observed yield outcomes. Our findings suggest that although climate anomalies significantly influence agricultural productivity, their effects are often modulated by farm-level management practices. This underscores the need for a nuanced interpretation of climate-yield relationships. Additionally, we investigate how different extreme events, such as drought, extreme rainfall, heatwave, and frost, vary in their impacts on yields revealing that these effects are neither uniform nor predictable across all conditions. These four events were selected due to their significant impacts on crop productivity and their prevalence as recorded by climate monitoring institutions in New Zealand.  Finally, we explore alignments and discrepancies between climate anomalies and yield data, identifying instances where severe climate conditions do not consistently correspond to reduced yields, and vice versa. These findings highlight the complexities inherent in extreme event's impact assessments and emphasise the importance of integrating climate data with local management and environmental factors for a robust understanding of yield variability.

The remainder of this paper is structured as follows. Section~\ref{sec:related_work} reviews relevant studies on anomaly and outlier detection methodologies with specific attention to agricultural and climate-related applications. Section~\ref{sec:data} describes the dataset used, including data sources, granularity, temporal scope, and detailed definitions of extreme events according to New Zealand’s national weather service, the National Institute of Water and Atmospheric Research (NIWA). The methodology of how we employ Isolation Forest and pre-processing techniques in this investigation is presented in Section~\ref{sec:methodology}. Results highlighting the alignment of detected extreme events with observed yield variations across farms are discussed in Section~\ref{sec:results}. Section~\ref{sec:diss} provides an in-depth discussion on the implications of our findings, limitations of current approaches, variability in the impact of extreme events, and proposed strategies for improvement. Finally, Section~\ref{sec:con} concludes the paper by summarising our key contributions and insights derived from this study.

\section{Related Work}
\label{sec:related_work}
% outlier detection on EE or Agri
Anomaly and outlier detection methods have been extensively adopted across diverse domains, including electricity monitoring \cite{oprea2021anomaly}, equipment fault diagnosis \cite{purarjomandlangrudi2014data}, and medical signal analysis \cite{baur2021autoencoders}, primarily due to their efficacy in identifying irregularities that signal potential system failures. In environmental sciences, such methodologies play a crucial role in detecting anomalous climate events. Flach et al. \cite{flach2017multivariate} compared multiple dimensionality reduction and anomaly detection algorithms, such as k-nearest neighbours mean distance, kernel density estimation, and recurrence analysis, and demonstrated their effectiveness in identifying environmental anomalies using Earth observation data. Mahmoud and Gan \cite{mahmoud2018urbanization} further advanced this domain by employing non-parametric change-point detection integrated with GIS-based decision support systems, assessing urbanisation and climate change impacts on flood risks in Egypt to produce detailed flood susceptibility maps. Expanding the methodological scope, Hael and Yuan \cite{hael2020identifying} proposed a functional outlier detection approach employing functional bag-plots and highest-density region box-plots, successfully identifying extreme rainfall anomalies in Yemen over a 21-year dataset. More recently, Zhang et al. \cite{zhang2022anomaly} introduced an approach based on Support Vector Data Description (SVDD), utilising multi-class classification and multi-label recognition techniques to effectively discriminate normal patterns from extreme events, including typhoons and earthquakes.

In agricultural applications, the adoptions of anomaly detection have been specifically tailored to address cultivation practices and crop health monitoring. Peichl et al. \cite{peichl2021machine} utilised random forest models to identify yield anomalies in winter wheat attributable to soil moisture variability and extreme weather events, highlighting soil moisture as having a more significant impact compared to meteorological variables, particularly when combined with spatial clustering techniques. Similarly, Mouret et al. \cite{mouret2021outlier} effectively employed Isolation Forests to analyse Sentinel-1 and Sentinel-2 time series data to detect anomalous crop development at the parcel level. Castillo-Villamor et al. \cite{castillo2021earth} developed an Earth Observation-based Anomaly Detection (EOAD) system that automated the thresholding of optical vegetation indices derived from Sentinel-2 and PlanetScope imagery achieving high accuracy in identifying yield-limiting anomalies in rice fields. Subsequently, Moso et al. \cite{moso2021anomaly} proposed an ensemble-based anomaly detection approach designed for smart agriculture data streams, successfully validating their approach on GPS logs from combine harvesters and crop datasets to enhance farm efficiency. Mujkic et al. \cite{mujkic2022anomaly} extended anomaly detection methodologies further into precision agriculture by leveraging convolutional autoencoders, particularly semi-supervised variants utilising max-margin loss, to detect unknown obstacles in autonomous agricultural vehicles. Most recently, Sjulg{\aa}rd et al. \cite{sjulgaard2023relationships} analysed historical temperature and precipitation anomalies across Sweden, highlighting significant yield losses resulting from extreme summer weather events, particularly affecting spring-sown crops due to their shorter growing periods and sensitivity to soil-texture interactions.

While these studies significantly contribute to the field of anomaly detection in environmental and agricultural contexts, they underscore the ongoing need for methods that can integrate advanced machine learning with explainable and interpretable frameworks in the real world. Addressing this gap could enhance stakeholder understanding and decision-making capability, providing practical solutions tailored for real-world agricultural and environmental management scenarios.

\section{Data}
\label{sec:data}
% data source, granularity, time scope, available features, and statistical table
\begin{figure}[!t]
\centering
\subfloat[]{\includegraphics[width=0.45\textwidth]{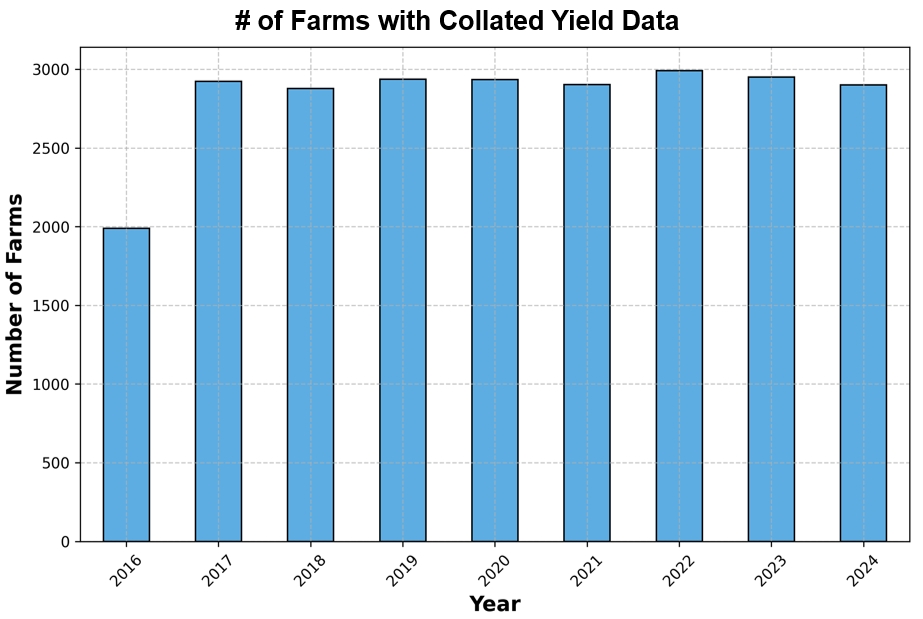}%
\label{farm_year}}
\hfil
\subfloat[]{\includegraphics[width=0.45\textwidth]{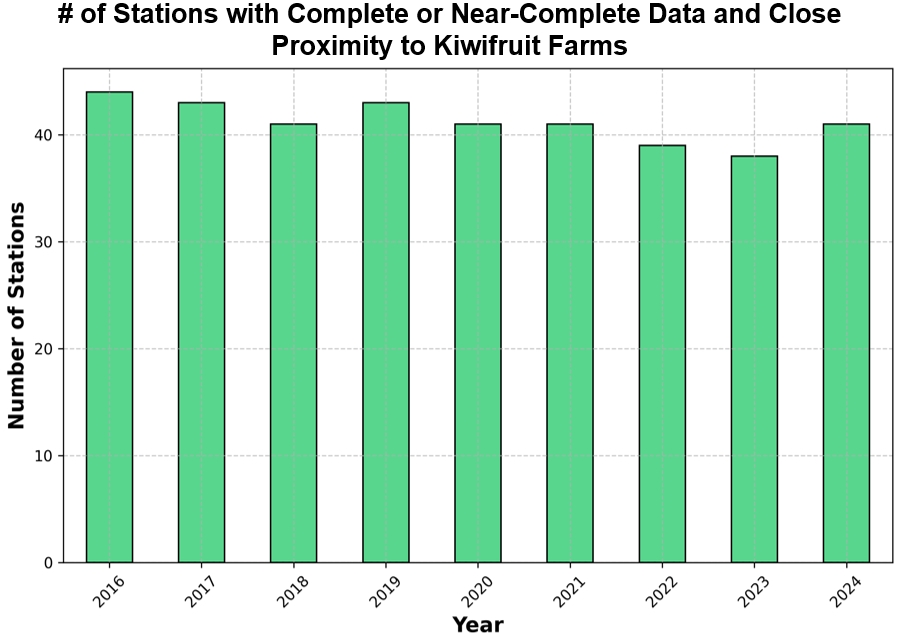}%
\label{station_year}}
\hfil
\subfloat[]{\includegraphics[width=0.45\textwidth]{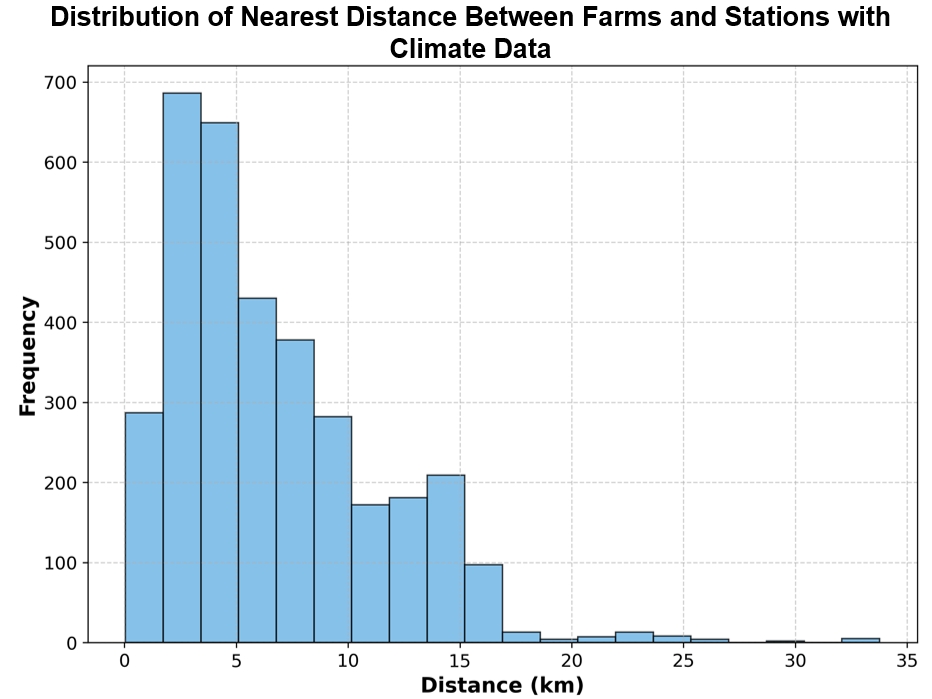}%
\label{station_farm_dis}}
\caption{Spatiotemporal statistics on the climate data and yield data.}
\label{station_farm}
\end{figure}

The datasets employed in this study comprise detailed climate and kiwifruit yield records from New Zealand, selected to facilitate an analysis of climate impacts on regional orchard productivity. Specifically, the kiwifruit yield data were sourced from The Yield Technology Solutions\footnote{\url{https://www.theyield.com/}} (now Yamaha Agriculture), covering a total of 3,437 farms, thus enabling high-resolution, farm-level analysis. Concurrently, climate observations were obtained through the NIWA National Climate Database (CliFlo database)\footnote{\url{https://niwa.co.nz/climate-and-weather/national-climate-database}}, providing daily meteorological records from a network of 318 climate stations strategically distributed nationwide. To accurately integrate climate observations with farm-specific yield data, spatial matching procedures were applied, whereby each farm was linked to the nearest available climate stations, typically within a 10 km radius, with very few cases extending beyond 15 km (as shown in Figure~\ref{station_farm_dis}). 
Temporally, the climate dataset spans approximately twelve years, from December 2008 to January 2021, although the completeness of records varies between individual stations. In contrast, kiwifruit yield data are available from 2016 to 2024, ensuring sufficient temporal overlap for comprehensive analysis of climate-yield relationships (see Figure~\ref{farm_year} and \ref{station_year}). The climate variables include daily maximum and minimum temperatures, solar radiation, precipitation, and wind parameters. However, the availability of these features varies considerably between stations, and inconsistencies in temporal coverage across the network pose analytical challenges, particularly for multivariate outlier detection methods.

\begin{figure}
    \centering
    \includegraphics[trim={3.5cm 0 2cm 0},clip,width=1.1\linewidth]{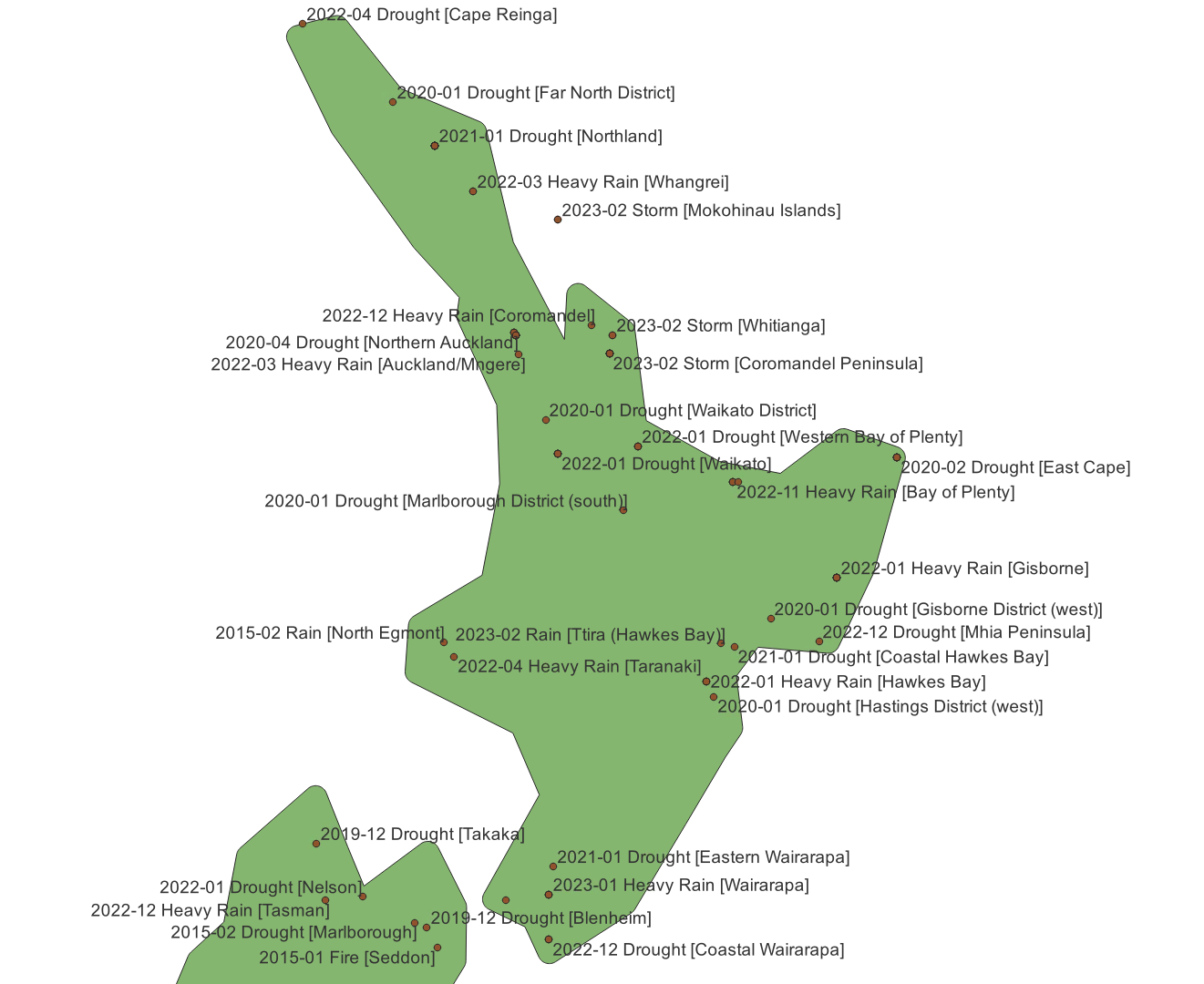}
    \caption{Samples of extreme events summarised from NIWA seasonal summary documents.}
    \label{fig:event_map}
\end{figure}

% EE records, and definitions from NIWA
Extreme events data were collected from multiple authoritative sources. Large-scale extreme events, including droughts and anomalies in temperature and rainfall, were extracted from NIWA seasonal summaries \cite{niwaseasonal}, which provide detailed regional documentation and enable precise identification and categorisation of event severity (see Figure~\ref{fig:event_map}). Severity assessments were established using rule-based categorisations relative to long-term climatic baselines. However, frost events, due to their localised nature and inconsistent reporting, are less comprehensively covered in NIWA summaries. To address this gap, supplementary frost event data were obtained from Zespri monthly orchard reports \cite{zesprimonthly}, although these records typically lack quantifiable severity information. Consequently, frost events were represented using a binary classification to indicate presence or absence. 

Collectively, the integration of these multi-source datasets aims to enhance the detection and characterisation of extreme weather events from daily climate data, thereby facilitating a robust assessment of their impacts on kiwifruit yields across regional orchards.

\section{Methodology}
\label{sec:methodology}

\begin{figure*}
    \centering
    \includegraphics[width=0.95\linewidth]{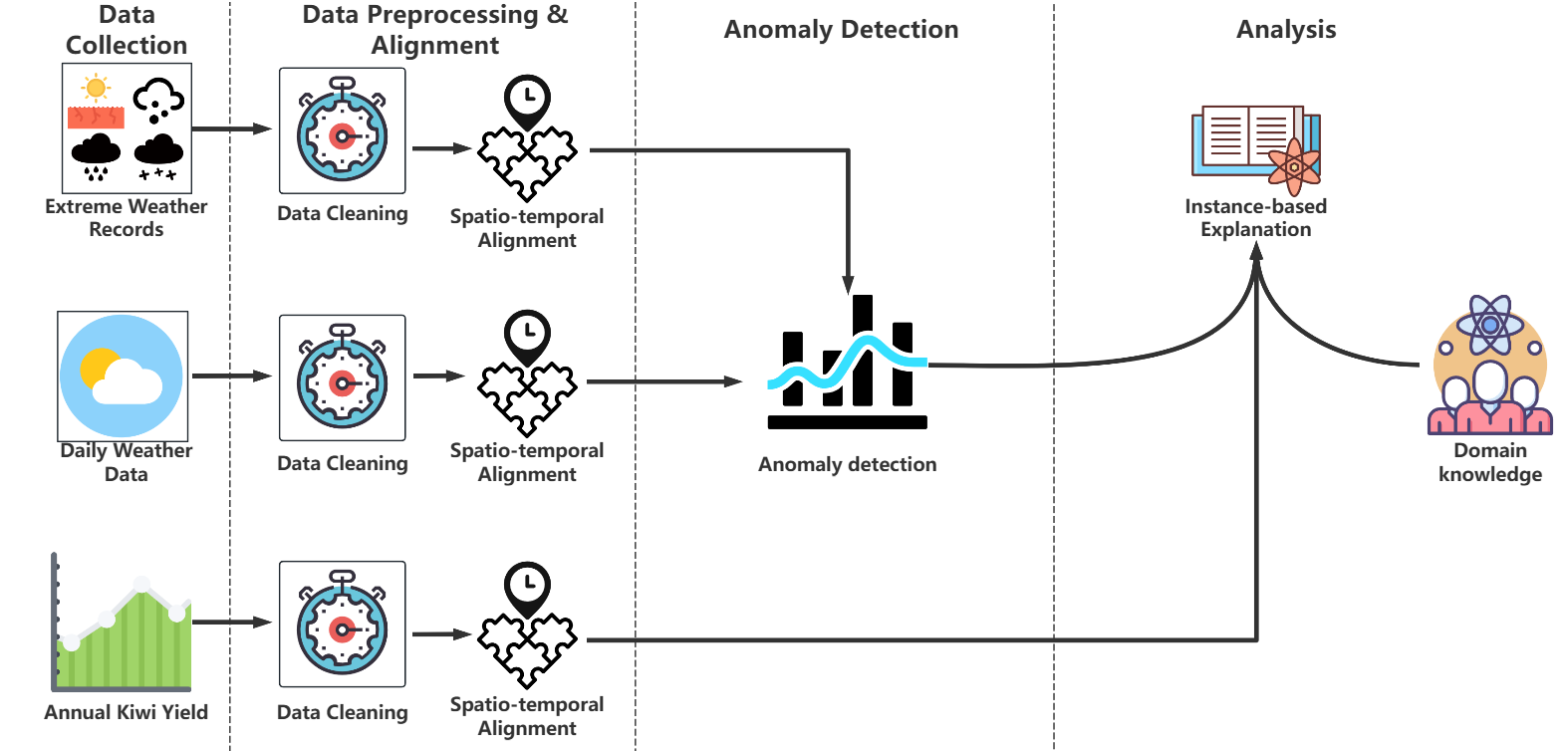}
    \caption{Overview of the methodology workflow. The approach integrates diverse data sources, pre-processing steps, Isolation Forest-based anomaly detection, and instance-based explanations supported by domain knowledge to evaluate the impact of extreme weather events on annual kiwifruit yield.}
    \label{fig:workflow}
\end{figure*}

% intro Isolated Forest
Figure~\ref{fig:workflow} presents our methodological workflow employed in this study. The analysis began with the acquisition of three aforementioned data sources. Each dataset underwent a structured pre-processing pipeline to ensure data integrity and consistency. The pre-processing phase commenced with a comprehensive data cleaning procedure, addressing inconsistencies, missing values (found to be only less than 10 days in between), and potential errors. Specifically, missing values were handled using forward filling, where each missing entry was replaced with the last observed value. Following data cleaning, all datasets were spatio-temporally aligned by matching observations to the nearest weather stations and synchronising data across overlapping years.

Subsequently, the processed datasets advanced to the anomaly detection phase. Isolation Forest (IF) was employed as the primary anomaly detection method due to its computational efficiency and its ability to effectively isolate anomalies through recursive partitioning. Unlike density- or distance-based approaches, IF leverages random feature-space partitioning to isolate anomalous points efficiently. Mathematically, IF constructs multiple isolation trees through recursive binary partitioning. Given a dataset $X = \{x_1, x_2, ..., x_n\}$, each tree partitions the feature space iteratively until all instances are isolated in leaf nodes. The anomaly score $s(x)$ of an instance $x$ is computed as:
\[
s(x, n) = 2^{-\frac{E(h(x))}{c(n)}}
\]
where $E(h(x))$ represents the average path length from the root node to the node containing $x$ across all isolation trees, and $c(n)$ is the expected path length of unsuccessful searches in a binary search tree, expressed as:
\[
c(n) = 2H(n - 1) - \frac{2(n - 1)}{n},
\]
where $H(i)$ denotes the harmonic number, defined as $H(i) = \ln(i) + 0.5772156649$ (Euler-Mascheroni constant). Instances exhibiting shorter average path lengths were classified as anomalies, as they were more readily isolated. The contamination hyperparameter of IF, which determines the proportion of data points identified as anomalies, was tuned between 0.005 and 0.05 through an iterative trial-and-error approach for each climate feature independently. The detected anomalies were subsequently validated against historical observational records to assess the model's precision and its capability in identifying significant deviations in climate data.

Finally, in the analysis phase, the results of the anomaly detection process were interpreted using instance-based explanations, incorporating domain expertise to contextualise and substantiate the findings. Additionally, the impact of extreme weather events on kiwifruit yield was quantitatively assessed by computing the reduction in yield relative to the average of preceding years.

% any preprocessing or feature-engineering steps: normalization, aggregation.

\section{Results}
\label{sec:results}
This section presents our findings from the integrated analysis of climate and agricultural data, highlighting the impacts of key climatic events on kiwifruit yields across northern New Zealand. Using anomaly detection and yield variation assessments, we examine how drought, heatwave, extreme rainfall, and frost influence crop production at a regional scale. The results provide insights into the extent of yield fluctuations under different climatic conditions, revealing both vulnerabilities to various extreme events and adaptive responses across farms and kiwifruit varieties: i) Green/Hayward (HW) variety, and ii) Gold (GA) variety.  Note that the farm identifiers (4-digit numbers) were replaced by unique capital letters in this paper due to business confidentiality.  The farms were selected due to their proximity to the named weather stations in northern New Zealand.

\subsection{Drought}
\begin{figure*}
    \centering
    \includegraphics[width=1\linewidth]{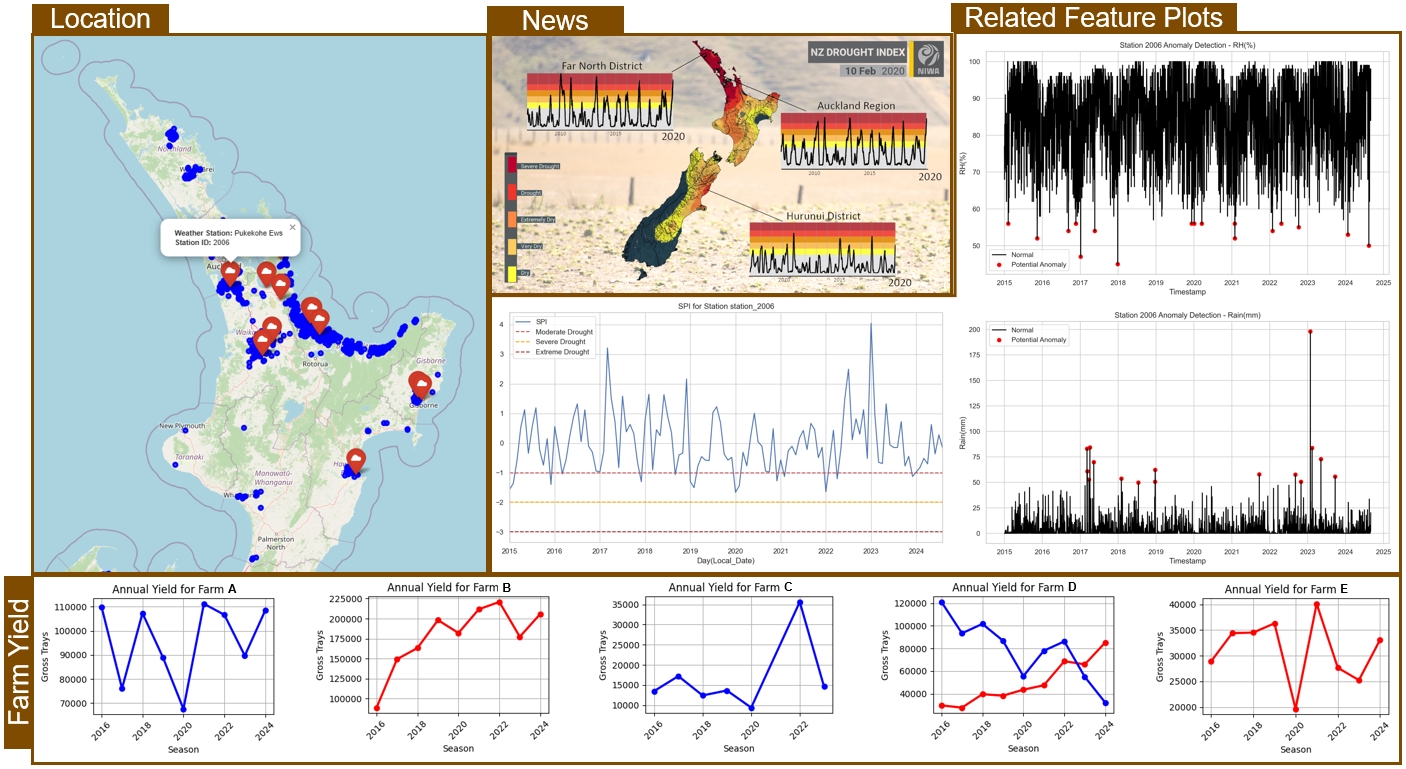}
    \caption{Weather station (red cloud icons) and kiwifruit farm locations (blue dots) in northern New Zealand, alongside climate anomalies, drought indicators, and annual kiwifruit yields. The panels illustrate drought severity (NZ Drought Index, February 2020), Standardised Precipitation Index (SPI) drought intensity (station-2006), anomalies detected in relative humidity and rainfall (red points), and yield patterns for two kiwifruit crop types (GA in red; HW in blue) from five selected farms.}
    \label{fig:drought}
\end{figure*}

The integrated climate and agricultural data presented in Figure~\ref{fig:drought} illustrate critical relationships between drought conditions, climate anomalies, and kiwifruit yield fluctuations across northern New Zealand. Several moderate to severe drought episodes occurred from 2015 to 2024, notably aligning with the significant drought event of 2020 reported by NIWA. Anomaly detection results for relative humidity and rainfall at station-2006 further validate these climatic irregularities. Correspondingly, annual kiwifruit yields exhibited substantial variability across analysed farms during the drought year. Yield losses ranged from approximately -28\% (increasing yield) to over 44\% compared to the previous 5-year average, with farm D (HW variety) experiencing the most severe reduction of 44\%, followed closely by farm E (GA variety) at approximately 41\%. Conversely, farm D (GA variety) and farm B (GA variety) showed a yield increase of about 29\% and 21\% respectively, reflecting variability even within the same location across different crop types. Farms cultivating the HW variety, such as farm A and D, also displayed significant yield reductions (around 29\% and 34\%, respectively). These findings highlight considerable variability in kiwifruit yields in response to drought conditions, emphasising the necessity for farm and variety-specific adaptation strategies.

% A 1556    B 2656  C 3996  D 4850  E 7267

\subsection{Heatwave}
\begin{figure*}
    \centering
    \includegraphics[width=1\linewidth]{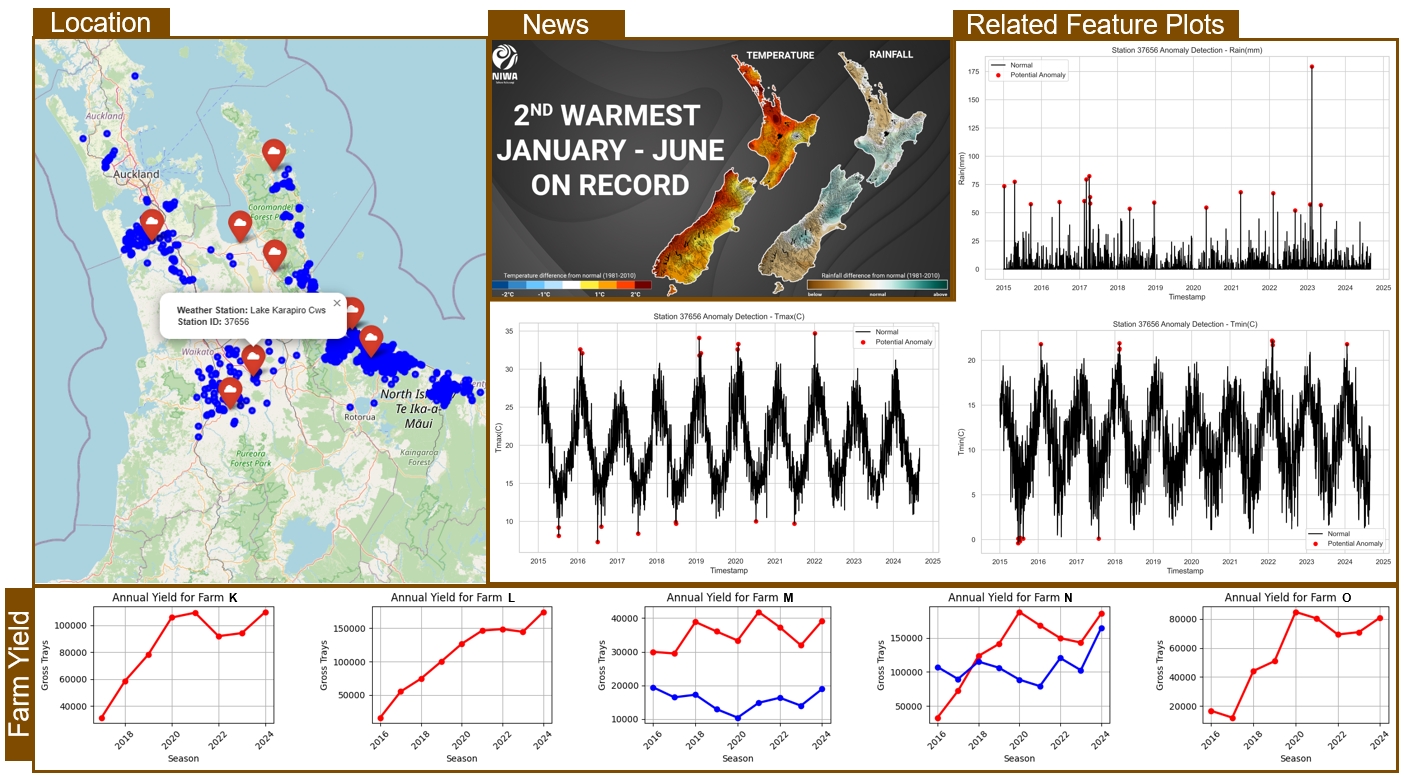}
    \caption{Locations of weather stations (red cloud icons) and kiwifruit farms (blue dots) in northern New Zealand, alongside climate anomalies related to temperature and rainfall, with anomaly detection results at station-37656 for maximum temperature (Tmax), minimum temperature (Tmin), and rainfall. Annual kiwifruit yields for two crop types (GA in red; HW in blue) from five farms are also illustrated.}
    \label{fig:heatwave}
\end{figure*}

The integrated climate and agricultural data presented in Figure~\ref{fig:heatwave} illustrate critical relationships between the 2022 heatwave event, climate anomalies, and kiwifruit yield fluctuations across northern New Zealand. Anomaly detection results at station-37656 clearly identify significant temperature deviations (Tmax and Tmin), confirming the presence of an exceptional heatwave, aligned with NIWA's report of record warmth from January to June 2022. Rainfall anomalies were comparatively less frequent during the same period. Interestingly, the yield data from selected farms indicate a generally positive influence from the heatwave, contrasting notably with drought-induced yield reductions observed previously. Compared to the 5-year window average from 2017 to 2021, all five selected farms demonstrated yield increment ranging from approximately 3\% to 47\%, notably farm H (GA) with an 3\% improvement, farm G (GA) demonstrated a remarkable yield increase of 47\%. Additionally, farm H (HW) and farm I (HW) also experienced modest yield improvements of approximately 13\% and 25\%, respectively. These positive responses highlight outstanding resilience for both kiwifruit varieties under heatwave conditions, reinforcing the various impact of extreme events on kiwifruit industry.
% F 1690    G 2083  H 3120  I 6674  J 7449

\subsection{Rainfall}
\begin{figure*}
    \centering
    \includegraphics[width=1\linewidth]{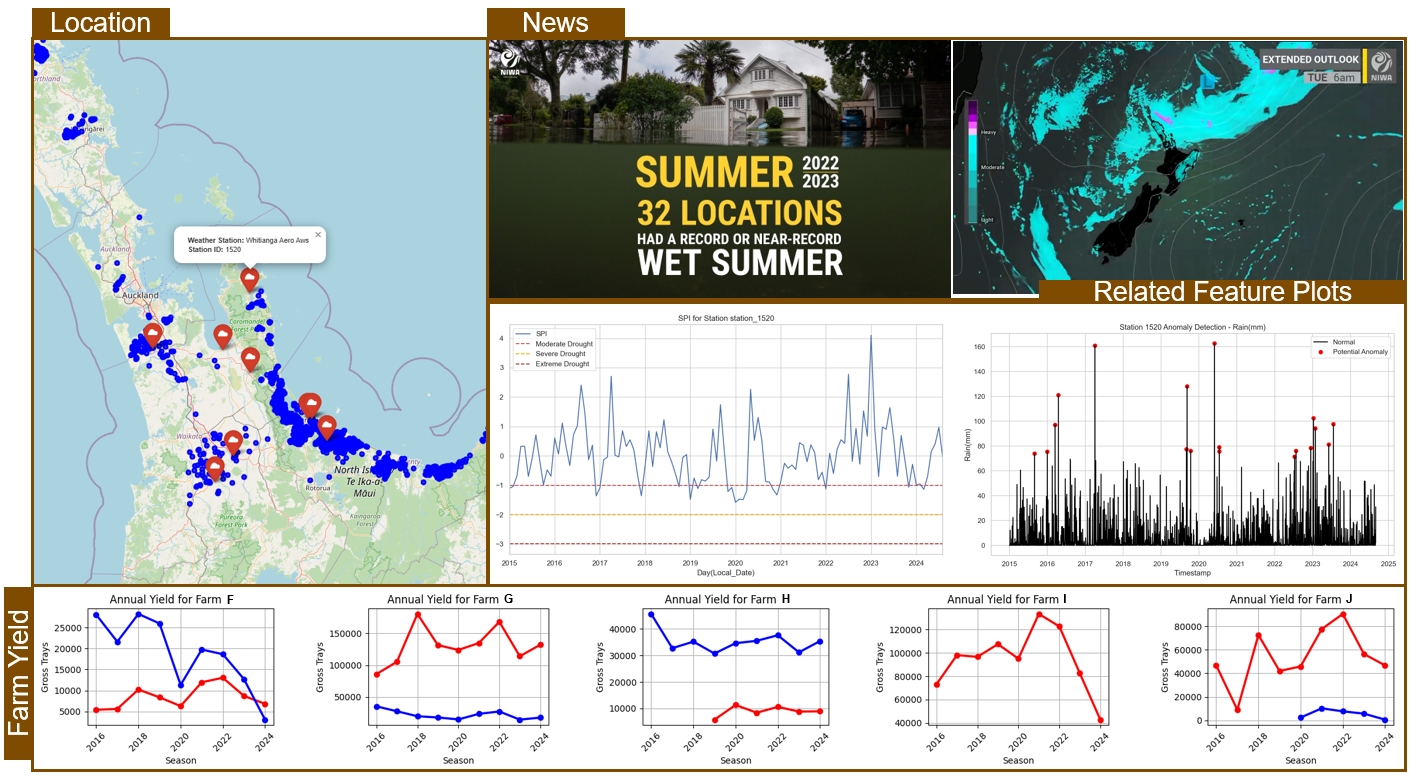}
    \caption{Locations of weather stations (red cloud icons) and kiwifruit farms (blue dots) in northern New Zealand, alongside evidence of record-breaking rainfall events during the summer of 2022-2023. SPI values and rainfall anomaly detection at station-1520 validate the extreme intensity of these events. Kiwifruit yield fluctuations for two crop types (GA in red; HW in blue) from five farms illustrate yield changes associated with extreme rainfall.}
    \label{fig:rainfall}
\end{figure*}

The integrated climate and agricultural data presented in Figure~\ref{fig:rainfall} illustrate significant impacts of extreme rainfall events on kiwifruit yields across northern New Zealand, particularly during the summer of 2022-2023. Anomaly detection results for rainfall at station-1520 confirm frequent high-intensity rainfall events, supported by the Standardised Precipitation Index (SPI)\footnote{\url{https://niwa.co.nz/nz-drought-indicator-products-and-information/drought-indicator-maps/standardised-precipitation-index-spi}} peaks and NIWA's reports identifying record or near-record wet conditions at multiple locations. Correspondingly, yields across the analysed farms significantly declined in 2023, with losses ranging from approximately 3\% to as high as 38\%. Farm K and L (both HW variety) faced the most severe reduction of 38\% and 30\% respectively, while farm M (GA variety) experienced the least yield decrease (approximately 3\%). Notably, both kiwifruit varieties, i.e., GA and HW, showed significant vulnerabilities, although variability within and between varieties persisted (GA: 21.27\%; HW: 22.83\%). These results clearly underscore the severe impact of extreme rainfall events on kiwifruit production, highlighting considerable risks and the critical need for farm-level adaptation and robust event management strategies.
% K 1023    L 5002  M 6462  N 8952  O 9832

\subsection{Frost}
\begin{figure*}
    \centering
    \includegraphics[width=1\linewidth]{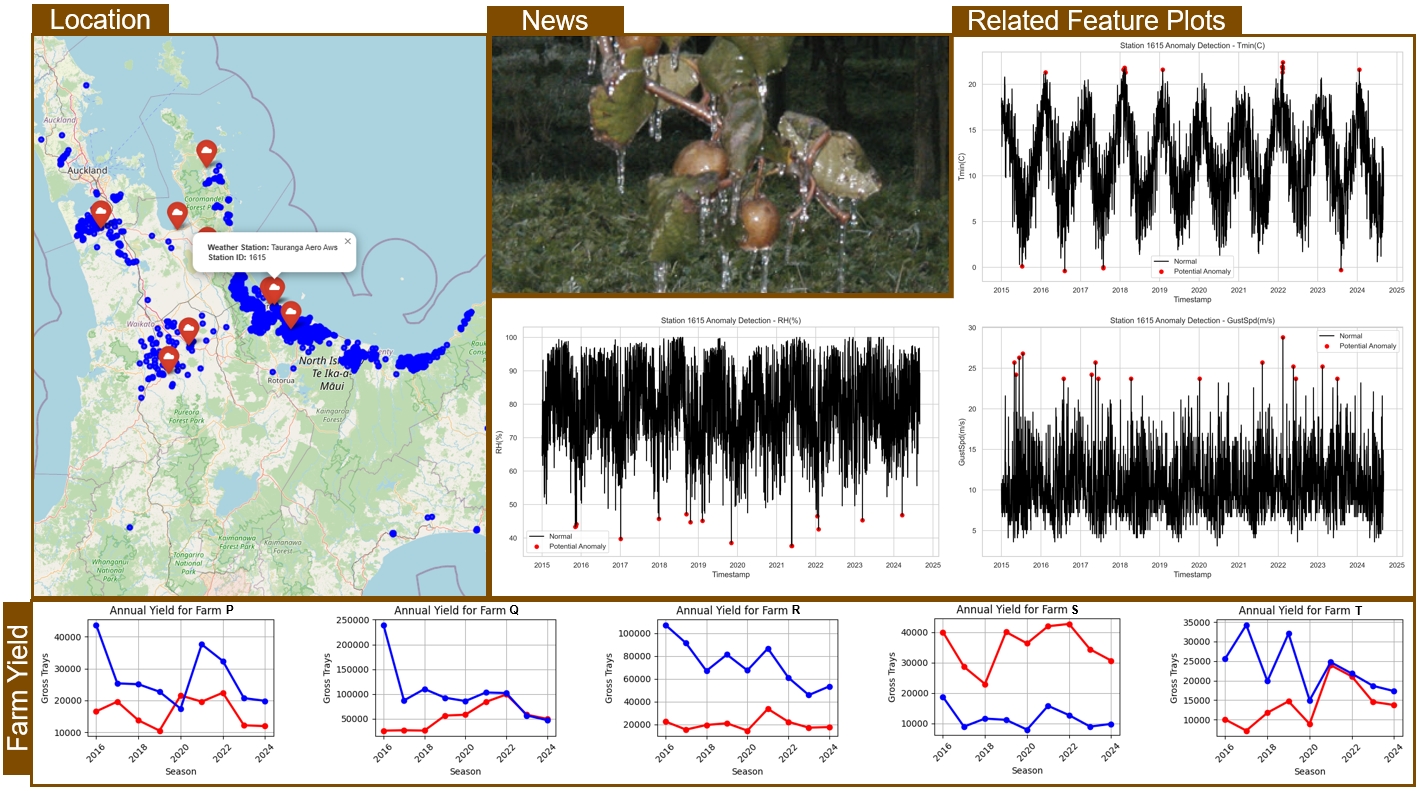}
    \caption{Locations of weather stations (red cloud icons) and kiwifruit farms (blue dots) in northern New Zealand, accompanied by evidence of frost occurrence in early 2023 (source: Zespri). Anomaly detection results at station-1615 for relative humidity (RH), minimum temperature (Tmin), and wind speed (GustSpd) provide further context. Yield data from five farms for two kiwifruit crop types (GA in red; HW in blue) demonstrate impacts associated with the frost event.}
    \label{fig:frost}
\end{figure*}

The integrated climate and agricultural data presented in Figure~\ref{fig:frost} illustrate substantial kiwifruit yield impacts resulting from frost events in early 2023 across northern New Zealand. Frost, particularly late-spring frost, tends to be localised and often overlooked by government climate agencies due to its small spatial scale, making detection from standard anomaly results challenging. Relative humidity (RH), minimum temperature (Tmin), and wind speed (GustSpd) are critical indicators included here due to their direct relevance to frost formation: frost typically occurs under conditions of low minimum temperatures, calm winds, and high relative humidity. Although anomaly detection at station-1615 identifies sporadic anomalies in these variables, the precise timing and localised nature of frost mean these broader-scale detections may not fully capture its true intensity and frequency. Nonetheless, yield data confirm significant agricultural impacts, with yield reductions ranging from approximately 7\% to over 42\%. Farm P (26.17\%), farm Q (29.67\%) and farm R (33.66\%) experienced the most severe yield losses on both GA and HW, emphasising kiwifruit vulnerability to frost. Conversely, farm S (GA) experienced the smallest reduction, approximately 7\%. These substantial yet variable yield impacts underscore frost as a critical, though frequently underestimated, threat to kiwifruit productivity. A comparative analysis of yield losses between GA (13.33\%) and HW (35.22\%), conducted on five randomly selected farms, indicates that GA exhibits superior resilience to frost conditions.
% P 1086    Q 4300  R 6637  S 7494  T 8456

\section{Discussion and Future Directions}
\label{sec:diss}
\begin{figure*}
    \centering
    \includegraphics[width=0.8\linewidth]{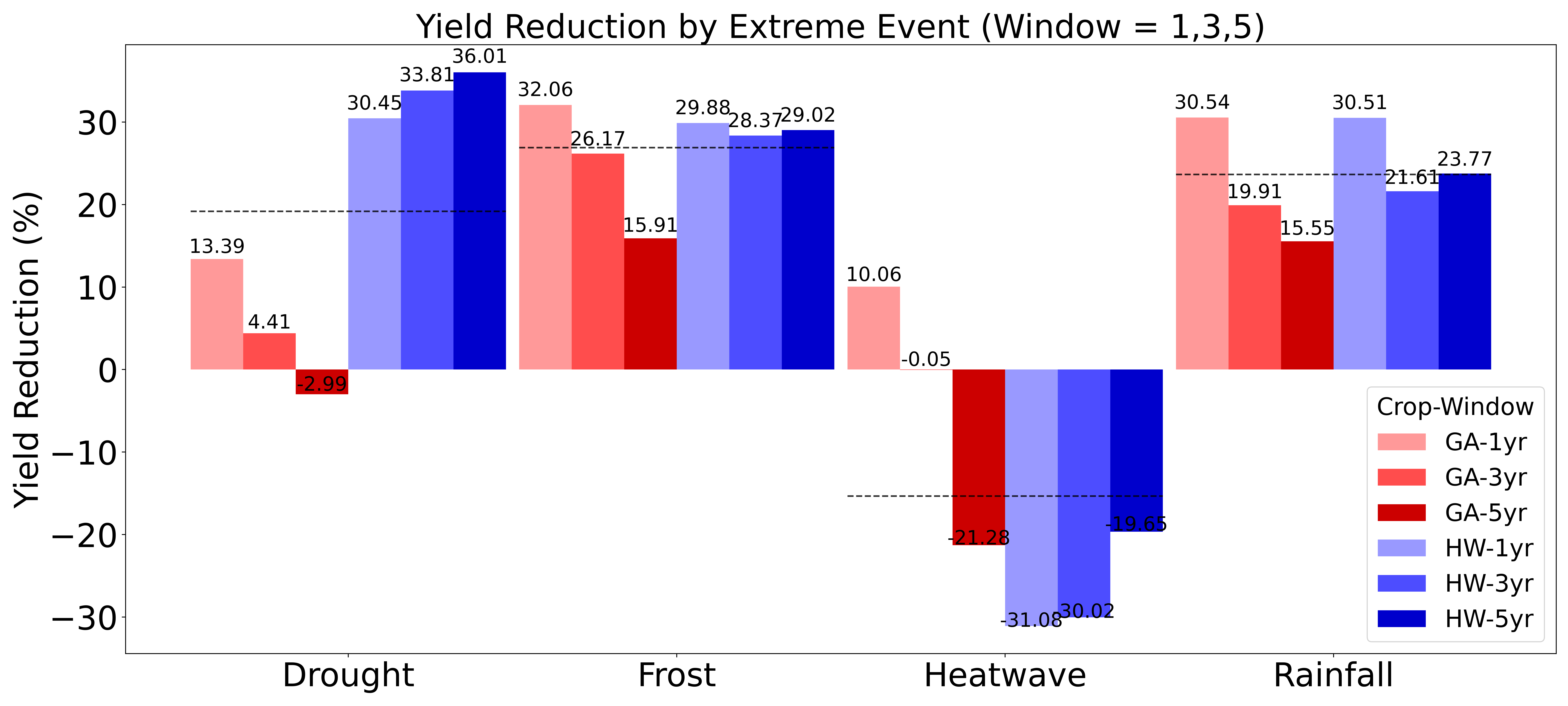}
    \caption{Yield reduction percentages under different window averaging sizes. Each group consists of six bars, with the horizontal dashed lines indicating the mean yield reduction within each group.}
    \label{fig:window}
\end{figure*}

The findings presented above reveal significant variability in the effects of extreme events, namely drought, extreme rainfall, heatwave, and frost, on kiwifruit yield in northern New Zealand, as shown in Figure~\ref{fig:window}. Among these, frost events exhibited the most severe impact, with average yield losses of approximately 27\% across the five analysed farms. This was followed by extreme rainfall at 23\%, drought at 20\%, and heatwave, which had the least adverse effect with average benefit around 15\%. The pronounced impact of frost highlights a major but often under-recognised threat to kiwifruit production, likely attributable to the localised and transient nature of frost events, which may lead to their under-reporting. On the other hand, when yield changes were disaggregated by kiwifruit variety, further patterns emerged. Both GA and HW varieties experienced substantial reductions during drought conditions, though intra-group variability was evident. On average, HW kiwifruit suffered greater losses (33\%) compared to GA (5\%). Heatwave, however, produced mixed outcomes: GA farms even reported yield increases (up to 31\%), while HW farms showed around average yield with previous years. Extreme rainfall in early 2023 resulted in significant yield reductions for both types -- approximately 22\% for GA and 25\% for HW. Similarly, frost caused major losses across both varieties, with average reductions of 25\% for GA and 29\% for HW, underscoring the widespread vulnerability of kiwifruit to frost and excessive rainfall, regardless of cultivar.
% U 1449 V 8134   W 8816   X 4563   Y 6674   Z 8152   AA 2520  AB 6547  AC 8900
\begin{figure*}
    \centering
    \includegraphics[width=0.9\linewidth]{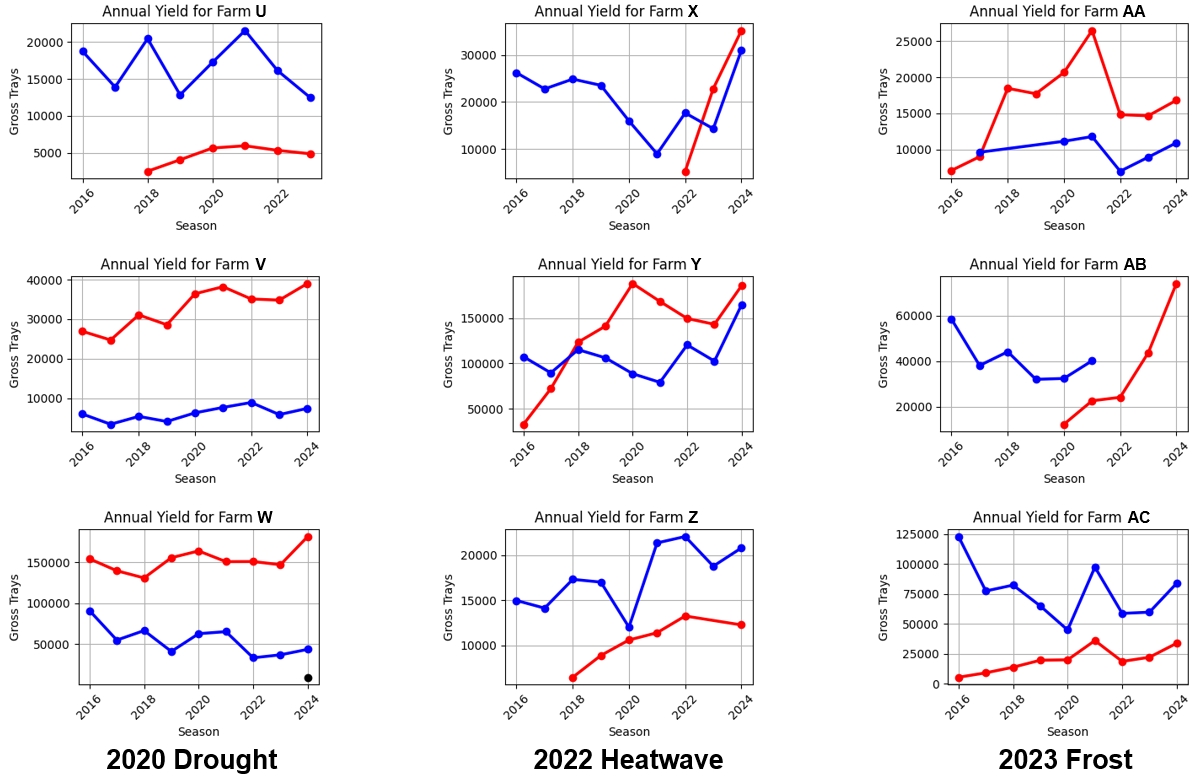}
    \caption{Counterexamples of kiwifruit yield for 2020's drought, 2022's heatwave, and 2023's frost. For extreme rainfall, we find zero counterexample (GA in red; HW in blue).}
    \label{fig:count_example}
\end{figure*}

While anomaly detection techniques effectively identified broader climatic anomalies such as drought and extreme rainfall, they showed limitations in capturing the highly localised and transient nature of frost events. The challenge in detecting frost through standard anomaly indicators (e.g., relative humidity, minimum temperature, wind speed) highlights the limitations of generalised anomaly detection methodologies and points to the need for finer-scale, farm-specific monitoring systems. Apart from the limitation from the algorithm itself, the limited record of frost events in official reports also obstacle the research on frost in New Zealand. Based on the existing reports from kiwifruit industry \cite{zesprifrost}, they suggest significant variability in frost impacts among growers. The reported variability in yield reduction aligns with the observed wide range in farm-level data within this study. Such variability may contribute to the absence of comprehensive governmental reporting on frost impacts.

Additionally, yield analyses revealed several counterexamples where farms experienced limited impacts despite exposure to extreme climatic events. Figure~\ref{fig:count_example} illustrates orchards that exhibited relatively minor yield reductions. Based on empirical observations, identifying such counterexamples was more feasible for drought events than for extreme rainfall or frost, suggesting that drought impacts may be more amenable to mitigation through human intervention. In contrast, orchards appear to be less resilient to intense rainfall and frost, which are more difficult to manage effectively at the farm level. These insights also underscore the importance of incorporating farm-specific management variables (such as canopy architecture, irrigation strategies, and frost protection techniques) into future modelling efforts aimed at explaining or forecasting yield variation.
Moreover, an important but under-explored dimension is the additive effect of multiple extreme events occurring within a single season. Situations where drought, intense rainfall, and frost co-occur --- either sequentially or in combination --- may compound stress on crop systems in non-linear ways. Understanding these compound events is crucial for both modelling and intervention, as their combined impact may not be adequately captured by analysing each %stressor
of them in isolation.
To address spatial variability in yield outcomes across different stations and farms, we also recommend the application of ensemble-based analyses. Integrating data from multiple weather stations and farms can help smooth out localised anomalies and counterexamples, leading to more robust and generalisable inferences. Ultimately, such approaches will enhance the accuracy of future yield predictions and support the development of more effective and site-specific adaptation strategies.

\section{Acknowledgements}
This research was supported by the Food Agility Cooperative Research Centre under the project Yield Prediction Explainability \& Climate Adaptiveness (Project Code: FA128), a collaboration between Food Agility CRC Limited, The Yield Technology Solutions Pty Ltd (Yamaha Agriculture), and the Data Science Institute at the University of Technology Sydney. The authors gratefully acknowledge this support, which made the research and its dissemination possible.

\section{Conclusion}
\label{sec:con}
This study examined the impact of extreme climate events on New Zealand's kiwifruit yield using anomaly detection techniques. Our findings highlight significant yield reductions due to extreme rainfall and frost, while heatwave showed mixed effects. The results underscore the complexity of climate-yield interactions. Additionally, limitations in detecting localised events like frost suggest the need for finer-scale meteorological data and improved detection methods. Future work should integrate multi-farm data and agronomic variables to enhance anomaly detection and yield prediction supporting more resilient agricultural practices.
%\newpage
\bibliographystyle{IEEEtran}
\bibliography{reference}
\end{document}